\documentclass{INTERSPEECH2023}

\interspeechcameraready{}

\usepackage{graphicx}
\graphicspath{ {./images/} }

\newcommand{\numdralpairs}{1871}

\title{Towards cross-language prosody transfer for dialog}
\name{Jonathan E. Avila\(^1\), Nigel G. Ward\(^1\)}
\address{
  \(^1\)University of Texas at El Paso, United States}
\email{jonathan.edav@gmail.com, nigelward@acm.org}

\begin{document}

\maketitle

\begin{abstract}
  Speech-to-speech translation systems today do not adequately support use for dialog
  purposes. In particular, nuances of speaker intent and stance can be lost due to
  improper prosody transfer. We present an exploration of what needs to be done to
  overcome this. First, we developed a data collection protocol in which bilingual
  speakers re-enact utterances from an earlier conversation in their other language, and
  used this to collect an English-Spanish corpus, so far comprising 1871 matched
  utterance pairs. Second, we developed a simple prosodic dissimilarity metric based on
  Euclidean distance over a broad set of prosodic features. We then used these to
  investigate cross-language prosodic differences, measure the likely utility of three
  simple baseline models, and identify phenomena which will require more powerful
  modeling. Our findings should inform future research on cross-language prosody and the
  design of speech-to-speech translation systems capable of effective prosody transfer.
\end{abstract}
\noindent\textbf{Index Terms}: speech-to-speech translation, corpus, prosodic dissimilarity metric, English, Spanish

\section{Introduction}

Speech-to-speech translation systems are valuable tools for enabling cross-language
communication. While very useful today for short, transactional interactions, they are
less so for long-form conversation~\cite{liebling2020}. One reason is that, without
proper prosody transfer, translation systems are unable to reliably convey many intents
and stances, impeding users' ability to deepen their interpersonal relationships and
social inclusion. In dialog, prosody conveys pragmatic functions such as in
turn-taking, expressions of attitudes, and negotiating agreement. Regarding prosody,
current translation systems generally aim only to produce prosody that sounds natural,
but this is not always sufficient.

In traditional models, translation is done by a cascade of subsystems --- for automatic
speech recognition, machine translation, and speech synthesis --- and the intermediate
representations are just text, with all prosodic information lost. The prospect instead
of transferring the additional information provided by the source-language prosody was
a motivation for the development of unified, end-to-end models~\cite{jia2022}. Despite
rapid recent
advances~\cite{popuri2022,lee2022,lee2022textless,dong22b_interspeech,jia22b_interspeech},
the ability of such models to perform prosody transfer seems not to have been examined.
Rather, current approaches to prosody transfer handle it with specific
modules~\cite{do2015improving,kano2012method,huang2023}. To date, these target only
specific functions of prosody, notably its roles in conveying paralinguistic/emotional
state, emphasis, and syntactic structure, and target only a few prosodic features,
notably F\(_0\), pausing, and word duration. Very recent work has shown that this can
significantly improve perceived translation quality~\cite{huang2023}, but also that
these techniques so far only close less than half of the perceived gap between default
prosody and the human reference. Clearly, something is still missing. This paper
investigates what that might be.

While one might hope that the answer could be found in the linguistics literature,
published knowledge of how prosody differs across languages focuses mostly on
syllable-level, lexical, and syntactic prosody. In particular, there is relatively
little work on differences in how prosody conveys pragmatic functions. Even for English
and Spanish, a well-studied pair, our knowledge is sparse beyond a few topics such as
turn-taking~\cite{berry1994}, questions and
declaratives~\cite{farias2013,zarate2018production}, and expression of
certainty~\cite{ramirezverdugo2005}. However, these certainly do not exhaust the
prosodic meanings important for dialog. Further, these studies have been mostly limited
to differences in intonation and duration, leaving out most prosodic features.
Accordingly, this paper takes a fresh look, using a corpus-based approach.

\section{Protocol and corpus}

To investigate prosodic differences in dialog, we need a suitable cross-language
corpus. However, corpora for speech-to-speech translation today primarily comprise
monologues, derived from readings~\cite{wang2020a,ardila2020,pratap2020,boito2020},
political discussions~\cite{wang2021}, or informative
talks~\cite{salesky2021,cattoni2021,oktem2021}. Those comprising dialogs were derived
from television show dubs~\cite{oktem2021,huang2023}, lectures and press
conferences~\cite{doi2021}, or speech synthesis~\cite{jia2022b,zhang2020}. Speech
collected in these settings lacks interactivity, spontaneity, and most of the prosodic
variation found in real dialog.

We accordingly developed the Dialogs Re-enacted Across Languages (DRAL) protocol. This
involves pairs of nonprofessional, bilingual participants. They first have a ten-minute
conversation, which we record. These conversations are unscripted, although we
sometimes suggest topics, which allows for pragmatic diversity and spontaneous
interactions. Depending on their relationship, the participants mostly get to know each
other, catch up on recent happenings, and/or share personal experiences. Subsequently,
under the direction of a producer, they select an utterance or exchange and closely
re-enact it in their other language, which may take several attempts to get right. They
then re-enact another utterance. The yield is typically a few dozen matched pairs per
one-hour session, with overall good pragmatic diversity, as suggested by
Table~\ref{table:example-notes}. Our design choices and the DRAL corpus are discussed
further in our technical report~\cite{ward2022c}.

Following this protocol we have so far collected \numdralpairs{} matched EN-ES
utterance pairs, from a total of 42 speakers. The latest release, including source
recordings and metadata, is available at \url{https://cs.utep.edu/nigel/dral/}. In the
following explorations, we use the first 1139 matched ``short'' utterances, which each
feature a single interlocutor. The average duration is 2.5 seconds.

\section{Utterance prosody representation}\label{sec:prosody-representations}

As our aim here is exploratory, we chose to work with simple, explicit, interpretable
representations of prosody. We use the Midlevel Prosodic Features
Toolkit\footnote{\url{https://github.com/nigelgward/midlevel}}, as its features were
designed to be robust for dialog data, generally perceptually relevant, and normalized
per speaker. From the available features, we selected ten based on previous utility for
many tasks for several languages~\cite{ward2019}, specifically: intensity, lengthening,
creakiness, speaking rate, pitch highness, pitch lowness, pitch wideness, pitch
narrowness, peak disalignment (mostly late peak), and cepstral peak prominence smoothed
(CPPS), the latter an inverse proxy for breathy voice. This rich set of prosodic
features supports more comprehensive analyses than most prosody research efforts.

To characterize the prosody of an utterance, each base feature is computed over ten
non-overlapping windows, together spanning the whole utterance. Thus, each utterance is
represented by 100 features. The window sizes are proportional to an utterance's
duration and span fixed percentages of its duration: 0--5\%, 5--10\%, 10--20\%,
20--30\%, 30--50\%, 50--70\%, 70--80\%, 80--90\%, 90--95\%, 95--100\%, as seen in
Figure~\ref{figure:intensity-cpps-example}. This representation is thus not aligned to
either syllables or words, but is appropriate for representing the sorts of overall
levels and contours that are most often associated with pragmatic functions.
Normalization occurs at two steps in the feature computation. The low-level
(frame-level) features --- pitch, energy, and CPPS --- are normalized per track to
mitigate individual differences. Subsequently, the mid-level features (peak
disalignment, lengthening, etc.) are computed over each specified span for every
utterance, and after being computed for all utterances in a track, each is
z-normalized.

\section{Cross-language feature correlations}

For our first glimpse at the EN-ES prosody mapping, we examined the Spearman
correlations between the 100 EN prosodic features and the 100 ES prosodic features,
across all matched pairs. (We computed Spearman correlations as well within each
language for comparison.) Were EN and ES prosodically identical, we would expect each
EN feature to correlate perfectly with its ES counterpart. In fact, the correlations
were far more modest but always positive and often substantial: more than half the
features sharing the base feature and span have correlation \(\rho \ge0.3\). Thus,
overall, EN and ES prosody is quite similar, and pitch highness is generally the most
similar, especially towards the middle of utterances (e.g.\ 30--50\%, \(\rho=0.59\)).
While some features, such as pitch highness, have much stronger span-for-span
correlations, other features, notably speaking rate, lengthening, and CPPS, have
correlations that are strong throughout the utterances. For example, speaking rate at
every span in an EN utterance correlates with speaking rate at every span in the
corresponding ES utterance. These findings are compatible with the idea that English
and Spanish prosody is overall roughly similar, but that the locations of local
prosodic events can vary, likely due to differences in word order and lexical accents.

\begin{figure}[h]
  \begin{center}
    \includegraphics[width=8cm]{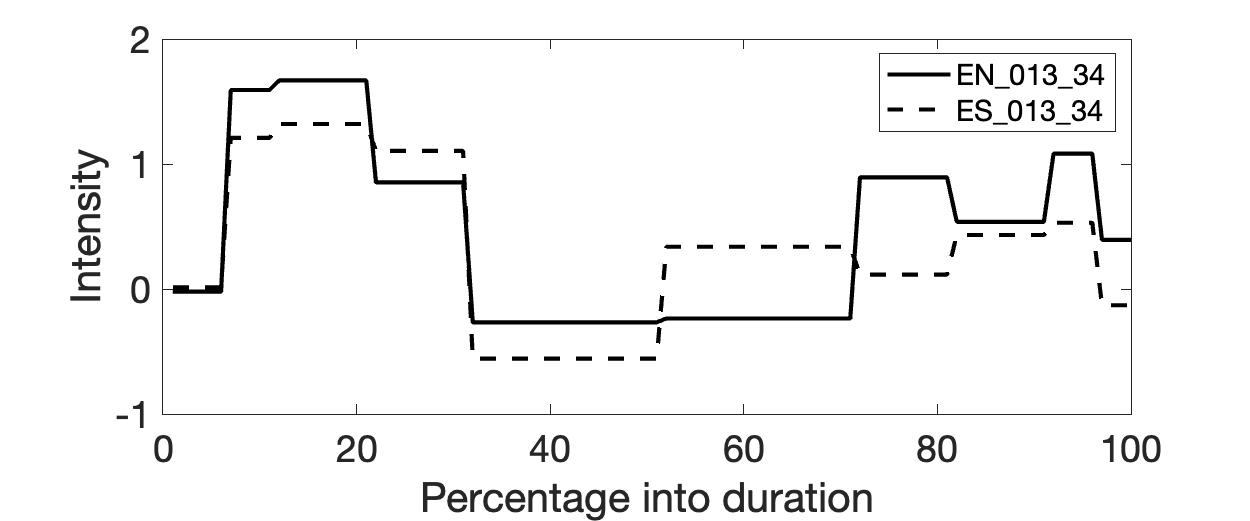}
    \includegraphics[width=8cm]{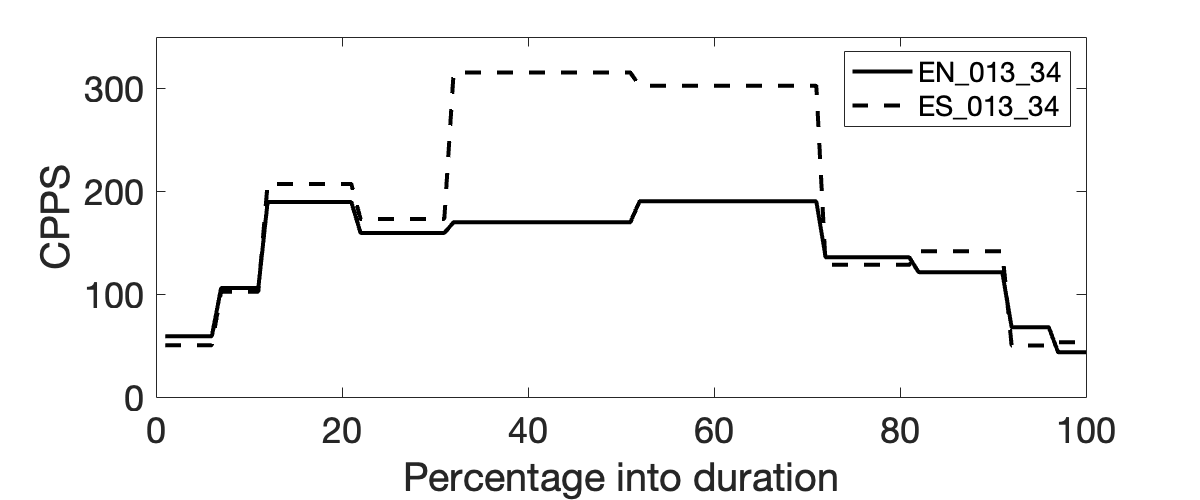}
    \caption{Example of a matched pair with EN high near final
      intensity and ES high CPPS\@. EN:``If you have an undergrad in
      anything, you can just, skip to a Master's in anything else''
      ES:\@``Si tienes carrera en cualquier cosa, puedes brincar a la
      maestría en lo que sea''.}\label{figure:intensity-cpps-example}
  \end{center}
\end{figure}

However, some correlations were much weaker. The lowest cross-language correlations for
the same features were for creakiness and peak disalignment, suggesting that these are
likely to have different functions in the two languages. There were also many
off-diagonal correlations. Most of these were unsurprising, such as the
anticorrelations between the speaking rate and lengthening features, but not all. For
example, intensity at the end of an EN utterance correlates with CPPS throughout an ES
utterance (EN 90--95\% vs. ES 5--20\%, 30--70\%, and 80--100\%, \(\rho\ge0.3\)), while
no such relationship was found within either language. Examination of the ten pairs
that most closely reflect this pattern (EN high near final intensity and ES high CPPS),
showed that in half the speaker is preparing a follow-up explanation. Thus, we have
identified a pragmatic function that seems to be prosodically marked differently in EN
and ES.\@ Figure~\ref{figure:intensity-cpps-example} shows the values for these two
features for one such pair.

\section{Prosodic dissimilarity metric}\label{section:metric}

To judge the quality of prosody transfer, we need a measure of how far the predicted
prosody diverges from the observed prosody in the human reference translation. If there
existed a synthesizer capable of realizing arbitrary prosodic specifications, we could
just use it and then use human perceptions of the match between the synthesized and
reference speech. However, no existing synthesizer is capable of this, especially for
the rich set of prosodic features we are investigating here. Existing metrics for
estimating similarity from prosodic feature representations exist, such
as~\cite{mary2013} and~\cite{rilliard2011}, but these again are limited in the prosodic
features considered.

Accordingly, we propose a new simple metric. This estimates the dissimilarity of two
utterances as the Euclidean distance between their respective prosodic representations,
as computed in Section~\ref{sec:prosody-representations}, with all features given equal
weight.

\begin{table*}[ht]
  \begin{center}
    \caption{An anchor and its most similar and dissimilar utterances, as estimated by the prosodic dissimilarity metric.\label{table:example-notes}}
    \begin{tabular}{lll}
      \toprule
      Utterance   & Transcription                                                                      & Role                            \\
      \midrule
      EN\_016\_16 & I would be kind of scared to ask questions to the professor or\ldots               & anchor                          \\
      EN\_034\_20 & It's like, I would do meds, but in a lotion form.                                  & similar  \rule{0ex}{2.6ex}      \\
      EN\_018\_12 & What have been like, some challenges for you in your career?                       & similar                         \\
      EN\_025\_1  & So overall, what music do you prefer to listen to?                                 & similar                         \\
      EN\_025\_7  & So I have to pick music that I like, but also that people\ldots                    & similar                         \\
      EN\_011\_41 & And I really like Mejia because he is the one always like telling                  & dissimilar    \rule{0ex}{2.6ex} \\
                  & \hspace{3ex} me ``Hey, you should apply to this, you should apply to this''        &                                 \\
      EN\_024\_1  & So uh yesterday you were telling me about, like, a weird, like, experience         & dissimilar                      \\
                  & \hspace{3ex} you had with the cops in Mexico, right?                               &                                 \\
      EN\_021\_13 & And the beach is really strange because it's like a, you see, like the             & dissimilar                      \\
                  & \hspace{3ex}  beach is not like a straight line. It was like a doughnut.           &                                 \\
      EN\_019\_19 & But do you think that someone who hasn't seen a Marvel move can just               & dissimilar                      \\
                  & \hspace{3ex}  watch any movie? Or is there any specific movies they have to watch? &                                 \\
      \bottomrule
    \end{tabular}
  \end{center}
\end{table*}

We do not expect this metric to accurately match human perceptions, but we can hope
that it might be useful as a first-pass metric for judging prosodic dissimilarity. To
gauge this, we compared its outputs to our perceptions of a few dozen within-language
utterance pairs. To structure this process, we wrote software to randomly select an
utterance (the ``anchor'') from the data and retrieve the four most similar utterances
and four most dissimilar utterances according to the metric. Ideally, perhaps, we would
have made holistic judgments of the degree of prosodic similarity between each
sample-anchor pair, but, probably like most people, we lack this ability. Instead, we
repeatedly listened and identified whatever similarities and dissimilarities we could
note, taking 2 or 3 minutes per pair to do so. The most salient of these were always at
the level of pragmatic function, rather than prosodic features, but we considered this
unproblematic, as the ultimate aim of prosody transfer is pragmatic fidelity, not
prosodic fidelity. We did this process for seven anchors and eight comparisons
utterances each, all from the English half of the data.

We found, first, that the metric captures many aspects of pragmatic similarity ---
including speaker confidence, revisiting unpleasant experiences, discussing plans,
describing sequences of events, and describing personal feelings --- all of which were
generally also prosodically similar. Table~\ref{table:example-notes} shows one set of
utterances to illustrate. The prosody of this anchor utterance suggested that the topic
is personal feelings: a slow then fast then slow speaking rate, a pause, and occasional
use of creaky voice. Each of the utterances rated similar by the metric shared these
qualities, albeit to varying degrees.

Second, we noted that the similarities found were not generally lexically governed.
While some words and syntactic structures have characteristic prosody, and some of the
pairs considered similar by the metric shared lexical content, such as {\em music} in
the fourth and fifth examples in Table~\ref{table:example-notes}, generally prosodic
similarity seemed to be orthogonal to lexical similarity.

Third, we noted that the metric does not always appear to match perceptions. To try to
understand its limitations and what needs improving, we examined examples where our
judgments diverged most from the metric's estimates, namely four which the metric
judged very similar but sounded rather different to us, including EN\_025\_1 in
Table~\ref{table:example-notes}, and two which we felt had significant similarities but
which the metric judged very different, including EN\_024\_1 in
Table~\ref{table:example-notes}. Of these, two pairs had very salient nasality
differences, which our model does not capture, and sounded very different in terms of
pragmatic function, specifically relating to the presumption of common ground. For
three pairs the problem seemed to be differences in syllable-aligned pitch and energy
contours, which are not directly represented by our features. However, for 50 of the 56
pairs examined, our judgments aligned with those of the model.

Thus, while the metric needs improving, overall we deemed it likely to be useful. We
consider these findings also to be evidence that our prosody representation is
meaningful. Accordingly, below we rely on both for evaluating the quality of prosody
transfer, as a way to obtain insight.

\begin{table}[ht]
  \begin{center}
    \caption{Utterance pairs partitions, chosen to have roughly a 20/80 split and at most share one unique speaker.}\label{tab:splits}
    \begin{tabular}{l r r}
      \toprule
               & \# utterance pairs & \# unique speakers \\
      \midrule
      Training & 912                & 20                 \\
      Testing  & 227                & 7                  \\
      \bottomrule
    \end{tabular}
  \end{center}
\end{table}

\section{Comparison of modeling strategies}

Our corpus and metric enable the evaluation of different models of the cross-language
prosody mappings. The task is, given the prosody of an utterance in the source
language, to predict the prosody of its translation in the target language. The error
is the dissimilarity between the inferred prosody and the prosody of the human
re-enactment. We here report the results for models in both directions, EN\(\rightarrow
\)ES and ES\(\rightarrow \)EN, using the partition described in Table~\ref{tab:splits}.

The first model is intended to represent the best that can be achieved with a typical
cascaded speech-to-speech model, with a synthesizer that operates in ignorance of the
input-utterance prosody. Our implementation relies on the lookup of the human-generated
translation in the target language, to avoid the impact of ASR or MT errors. We use
Whisper~\cite{openai2023} to transcribe this to a word sequence with punctuation and
then use Coqui TTS\footnote{\url{https://github.com/coqui-ai/TTS}} to synthesize speech
from that transcription. To ensure a fair comparison, utterances incorrectly
transcribed were excluded from the data. Table~\ref{tab:splits} reflects the 252
excluded utterances. To judge the quality of each output, we compute a representation
of the prosody of the synthesized speech using the method of
Section~\ref{sec:prosody-representations}.

The second model predicts the prosody of the translation to be identical to the prosody
of the input: it trivially outputs the same representation. This ``naive'' model
embodies a strategy of directly transferring the input prosody.

The third model is trained by linear regression. Thus, each feature of the target
prosody representation is predicted as a linear function of the 100 features of the
input utterance.

\begin{table}[ht]
  \begin{center}
    \caption{Model average error for prosody translation tasks.}\label{table:task-results}
    \begin{tabular}
      {>{\raggedright\arraybackslash}p{2.2cm}
      >{\raggedleft\arraybackslash}p{2cm}
      >{\raggedleft\arraybackslash}p{2cm}
      }
      \toprule
      Model             & EN\(\rightarrow \)ES task & ES\(\rightarrow \)EN task \\
      \midrule
      Synthesizer       & 12.65                     & 12.32                     \\
      Naive             & 11.35                     & 11.35                     \\
      Linear regression & 9.23                      & 9.37                      \\
      \bottomrule
    \end{tabular}
  \end{center}
\end{table}

Table~\ref{table:task-results} shows the three models' overall average error. The
synthesizer baseline is outperformed by the naive baseline, suggesting that keeping the
same prosody in translation may be a reasonable basic strategy. The naive baseline is
in turn outperformed by the linear regression model, suggesting that even a simple
model can learn some aspects of the mapping between English and Spanish prosody.

While our simple linear model shows a benefit, its prediction error is still very high.
We think the likely factors include not only the existence of mappings too complex for
a linear model, but also the small size of the training data, the existence of free
variation implying a permissible margin of error for our metric, unmodeled dependencies
of target-language prosody on the source-utterance context and its lexical content, and
speaker-specific prosody behavior tendencies.

\section{Qualitative analysis}\label{sec:failure-analysis}

To better understand the challenges of cross-language prosody modeling, we examined
examples where the various models did well or poorly.

First, we examined the 16 examples in each direction whose synthesized prosody was
least similar to the human-produced target. The most common and salient differences
were: failure to lengthen vowels and vary the speaking rate for utterances where
speakers are thinking or expressing uncertainty or hesitation, failure to change pitch
at turn ends, and generally sounding read or rehearsed and thus unnatural for
conversational speech.

Next, we exampled the 16 pairs for which the naive model did worse, that is, the cases
where the English and Spanish prosody diverged most. Often there were salient
differences, in a few common patterns, such as ES utterances being creakier than the
English, EN but not ES utterances ending with rising pitch, and EN utterances being
breathier in some regions. The latter two may reflect the common use of uptalk in
English, that is to say, the use of breathy voice and rising pitch to establish common
ground regarding a referent~\cite{ward2022two}, a pattern rare in the Spanish dialect
of our corpus. In other cases there were no highly salient differences; presumably,
these had multiple smaller differences which added up to a big difference according to
the metric.

Next, we examined the examples where the linear regression model provided the most
improvement relative to the naive baseline; unsurprisingly these were often cases where
it corrected for the divergences mentioned above.

Finally, we examined the highest-magnitude coefficients of the linear model. Most were
unsurprising and reflected correlations noted above. However, among the top three,
there was a --.32 coefficient relating EN lengthening over 5\%--10\% to ES CPPS over
0\%--5\%. This may reflect the tendency for EN speakers to start turns with fast speech
(low lengthening) but not ES speakers~\cite{ward2017}, who perhaps tend instead to
start turns with more harmonic (higher CPPS) speech.

\section{Implications and future work}\label{sec:implications}

As we expected, these investigations indicate that effective cross-language transfer
will require attention to prosodic features beyond pitch and duration. These include at
least breathy voice, creaky voice, and intensity. We also found that the prosody of
some pragmatic functions, as they occur in dialog, differs in previously unsuspected
ways across languages. These include at least grounding, getting personal, leading into
something, and taking the turn. These findings suggest that well-designed prosody
transfer techniques will be important for effective speech-to-speech translation.
Finally, our results indicate that doing so has the potential to convey many more
pragmatic functions and intents that have been previously managed.

These investigations relied on a small corpus, a non-comprehensive prosody
representation, and a crude metric. The fact that these enabled us to obtain
interesting findings, is evidence for their utility. At the same time, all of these
need extensions and improvements, and doing so would enable future work to produce a
clearer and broader picture of what prosody is conveying in the two languages, how it
does it, and what the differences are.

In addition to such basic research, we envisage our findings informing the design of
speech-to-speech translation systems, potentially via two paths. In one path, for
end-to-end models, an improved version of our dissimilarity metric, properly extended
and tuned to model human perceptions, could serve as the loss function for training. In
the other path, for cascaded models, our analysis techniques could inform the design of
a specific prosody-transfer module, and inspire the development of synthesizers capable
of following a rich prosody specification and thereby conveying a wide range of
pragmatic functions. Given the unavoidable high cost and consequent low volume of
matched conversation data, either approach will mostly likely need to exploit
per-language or joint self-supervised training techniques.

We share all our data, code, and observations at our public repository:
\url{https://github.com/joneavila/DRAL}.

\section{Acknowledgements}

We thank Emilia Rivas for assistance with the data collection, Ann Lee, Benjamin
Peloquin, and Justine Kao for discussions, and UTEP URI for internal funding.

\bibliographystyle{IEEEtran}
\bibliography{paper}

\begin{thebibliography}{10}
\providecommand{\url}[1]{#1}
\csname url@samestyle\endcsname
\providecommand{\newblock}{\relax}
\providecommand{\bibinfo}[2]{#2}
\providecommand{\BIBentrySTDinterwordspacing}{\spaceskip=0pt\relax}
\providecommand{\BIBentryALTinterwordstretchfactor}{4}
\providecommand{\BIBentryALTinterwordspacing}{\spaceskip=\fontdimen2\font plus
\BIBentryALTinterwordstretchfactor\fontdimen3\font minus \fontdimen4\font\relax}
\providecommand{\BIBforeignlanguage}[2]{{%
\expandafter\ifx\csname l@#1\endcsname\relax
\typeout{** WARNING: IEEEtran.bst: No hyphenation pattern has been}%
\typeout{** loaded for the language `#1'. Using the pattern for}%
\typeout{** the default language instead.}%
\else
\language=\csname l@#1\endcsname
\fi
#2}}
\providecommand{\BIBdecl}{\relax}
\BIBdecl

\bibitem{liebling2020}
D.~J. Liebling, M.~Lahav, A.~Evans, A.~Donsbach, J.~Holbrook, B.~Smus, and L.~Boran, ``Unmet {{Needs}} and {{Opportunities}} for {{Mobile Translation AI}},'' in \emph{Proceedings of the 2020 {{CHI Conference}} on {{Human Factors}} in {{Computing Systems}}}.\hskip 1em plus 0.5em minus 0.4em\relax {Association for Computing Machinery}, 2020, pp. 1--13.

\bibitem{jia2022}
Y.~Jia, M.~T. Ramanovich, T.~Remez, and R.~Pomerantz, ``Translatotron 2: {{High-quality}} direct speech-to-speech translation with voice preservation,'' in \emph{Proceedings of the 39th {{International Conference}} on {{Machine Learning}}}, 2022, pp. 10\,120--10\,134.

\bibitem{popuri2022}
S.~Popuri, P.-J. Chen, C.~Wang, J.~Pino, Y.~Adi, J.~Gu, W.-N. Hsu, and A.~Lee, ``Enhanced {{Direct Speech-to-Speech Translation Using Self-supervised Pre-training}} and {{Data Augmentation}},'' \emph{{arXiv}}, no. arXiv:2204.02967, 2022.

\bibitem{lee2022}
A.~Lee, P.-J. Chen, C.~Wang, J.~Gu, S.~Popuri, X.~Ma, A.~Polyak, Y.~Adi, Q.~He, Y.~Tang, J.~Pino, and W.-N. Hsu, ``Direct {{Speech-to-Speech Translation With Discrete Units}},'' in \emph{Proceedings of the 60th {{Annual Meeting}} of the {{Association}} for {{Computational Linguistics}} ({{Volume}} 1: {{Long Papers}})}, 2022, pp. 3327--3339.

\bibitem{lee2022textless}
A.~Lee, H.~Gong, P.-A. Duquenne, H.~Schwenk, P.-J. Chen, C.~Wang, S.~Popuri, J.~Pino, J.~Gu, and W.-N. Hsu, ``Textless speech-to-speech translation on real data,'' in \emph{NAACL}, 2022.

\bibitem{dong22b_interspeech}
Q.~Dong, F.~Yue, T.~Ko, M.~Wang, Q.~Bai, and Y.~Zhang, ``{Leveraging Pseudo-labeled Data to Improve Direct Speech-to-Speech Translation},'' in \emph{Proc. Interspeech}, 2022, pp. 1781--1785.

\bibitem{jia22b_interspeech}
Y.~Jia, Y.~Ding, A.~Bapna, C.~Cherry, Y.~Zhang, A.~Conneau, and N.~Morioka, ``{Leveraging unsupervised and weakly-supervised data to improve direct speech-to-speech translation},'' in \emph{Proc. Interspeech}, 2022, pp. 1721--1725.

\bibitem{do2015improving}
Q.~T. Do, S.~Sakti, G.~Neubig, T.~Toda, and S.~Nakamura, ``Improving translation of emphasis with pause prediction in speech-to-speech translation systems.'' in \emph{IWSLT}, 2015.

\bibitem{kano2012method}
T.~Kano, S.~Sakti, S.~Takamichi, G.~Neubig, T.~Toda, and S.~Nakamura, ``A method for translation of paralinguistic information,'' in \emph{Proceedings of the 9th International Workshop on Spoken Language Translation}, 2012, pp. 158--163.

\bibitem{huang2023}
W.-C. Huang, B.~Peloquin, J.~Kao, C.~Wang, H.~Gong, E.~Salesky, Y.~Adi, A.~Lee, and P.-J. Chen, ``A {{Holistic Cascade System}}, {B}enchmark, and {{Human Evaluation Protocol}} for {{Expressive Speech-to-Speech Translation}},'' \emph{{arXiv}}, no. arXiv:2301.10606, 2023.

\bibitem{berry1994}
A.~Berry, ``Spanish and {{American Turn-Taking Styles}}: {{A Comparative Study}},'' {Education Resources Information Center}, Tech. Rep. ED398747, 1994.

\bibitem{farias2013}
M.~G.~V. Far{\'i}as, ``A comparative analysis of intonation between {{Spanish}} and {{English}} speakers in tag questions, wh-questions, inverted questions, and repetition questions,'' \emph{Revista Brasileira de Lingu\'istica Aplicada}, vol.~13, no.~4, pp. 1061--1083, 2013.

\bibitem{zarate2018production}
G.~Z{\'a}rate-S{\'a}ndez, ``Production of final boundary tones in declarative utterances by {E}nglish-speaking learners of {S}panish,'' in \emph{Proceedings of the 9th International Conference on Speech Prosody. International Speech Communication Association (ISCA) Online Archive}, 2018, pp. 927--31.

\bibitem{ramirezverdugo2005}
D.~Ram{\'i}rez~Verdugo, ``The nature and patterning of native and non-native intonation in the expression of certainty and uncertainty: {{Pragmatic}} effects,'' \emph{Journal of Pragmatics}, vol.~37, no.~12, pp. 2086--2115, 2005.

\bibitem{wang2020a}
C.~Wang, A.~Wu, J.~Gu, and J.~Pino, ``{{CoVoST}} 2 and {{Massively Multilingual Speech-to-Text Translation}},'' in \emph{Interspeech}, 2021, pp. 2247--2251.

\bibitem{ardila2020}
R.~Ardila, M.~Branson, K.~Davis, M.~Kohler, J.~Meyer, M.~Henretty, R.~Morais, L.~Saunders, F.~Tyers, and G.~Weber, ``Common {{Voice}}: {{A Massively-Multilingual Speech Corpus}},'' in \emph{Proceedings of the 12th {{Language Resources}} and {{Evaluation Conference}}}.\hskip 1em plus 0.5em minus 0.4em\relax {European Language Resources Association}, 2020, pp. 4218--4222.

\bibitem{pratap2020}
V.~Pratap, Q.~Xu, A.~Sriram, G.~Synnaeve, and R.~Collobert, ``{{MLS}}: {{A Large-Scale Multilingual Dataset}} for {{Speech Research}},'' in \emph{Proc. {{Interspeech}}}, 2020, pp. 2757--2761.

\bibitem{boito2020}
M.~Zanon~Boito, W.~Havard, M.~Garnerin, {\'E}.~Le~Ferrand, and L.~Besacier, ``{{MaSS}}: {{A Large}} and {{Clean Multilingual Corpus}} of {{Sentence-aligned Spoken Utterances Extracted}} from the {{Bible}},'' in \emph{Proceedings of the 12th {{Language Resources}} and {{Evaluation Conference}}}.\hskip 1em plus 0.5em minus 0.4em\relax {European Language Resources Association}, 2020, pp. 6486--6493.

\bibitem{wang2021}
C.~Wang, M.~Riviere, A.~Lee, A.~Wu, C.~Talnikar, D.~Haziza, M.~Williamson, J.~Pino, and E.~Dupoux, ``{{VoxPopuli}}: {{A Large-Scale Multilingual Speech Corpus}} for {{Representation Learning}}, {{Semi-Supervised Learning}} and {{Interpretation}},'' in \emph{Proceedings of the 59th {{Annual Meeting}} of the {{Association}} for {{Computational Linguistics}} and the 11th {{International Joint Conference}} on {{Natural Language Processing}} ({{Volume}} 1: {{Long Papers}})}.\hskip 1em plus 0.5em minus 0.4em\relax {Association for Computational Linguistics}, 2021, pp. 993--1003.

\bibitem{salesky2021}
E.~Salesky, M.~Wiesner, J.~Bremerman, R.~Cattoni, M.~Negri, M.~Turchi, D.~W. Oard, and M.~Post, ``The {{Multilingual TEDx Corpus}} for {{Speech Recognition}} and {{Translation}},'' in \emph{Interspeech 2021}.\hskip 1em plus 0.5em minus 0.4em\relax {ISCA}, 2021, pp. 3655--3659.

\bibitem{cattoni2021}
R.~Cattoni, M.~A. Di~Gangi, L.~Bentivogli, M.~Negri, and M.~Turchi, ``{{MuST-C}}: {{A}} multilingual corpus for end-to-end speech translation,'' in \emph{Proceedings of the 2019 {{Conference}} of the {{North American Chapter}} of the {{Association}} for {{Computational Linguistics}}: {{Human Language Technologies}}}, vol.~1.\hskip 1em plus 0.5em minus 0.4em\relax {Association for Computational Linguistics}, 2019, pp. 2012--2017.

\bibitem{oktem2021}
A.~{\"O}ktem, M.~Farr{\'u}s, and A.~Bonafonte, ``Corpora compilation for prosody-informed speech processing,'' \emph{Language Resources and Evaluation}, vol.~55, no.~4, pp. 925--946, 2021.

\bibitem{doi2021}
K.~Doi, K.~Sudoh, and S.~Nakamura, ``Large-{{Scale English-Japanese Simultaneous Interpretation Corpus}}: {{Construction}} and {{Analyses}} with {{Sentence-Aligned Data}},'' in \emph{Proceedings of the 18th {{International Conference}} on {{Spoken Language Translation}} ({{IWSLT}})}.\hskip 1em plus 0.5em minus 0.4em\relax {Association for Computational Linguistics}, 2021, pp. 226--235.

\bibitem{jia2022b}
Y.~Jia, M.~Tadmor~Ramanovich, Q.~Wang, and H.~Zen, ``{{CVSS Corpus}} and {{Massively Multilingual Speech-to-Speech Translation}},'' in \emph{Proceedings of the {{Thirteenth Language Resources}} and {{Evaluation Conference}}}.\hskip 1em plus 0.5em minus 0.4em\relax {European Language Resources Association}, 2022, pp. 6691--6703.

\bibitem{zhang2020}
C.~Zhang, X.~Tan, Y.~Ren, T.~Qin, K.~Zhang, and T.-Y. Liu, ``{{UWSpeech}}: {{Speech}} to {{Speech Translation}} for {{Unwritten Languages}},'' in \emph{Proceedings of the {{AAAI Conference}} on {{Artificial Intelligence}}}, vol.~16, 2021, pp. 14\,319--14\,327.

\bibitem{ward2022c}
N.~G. Ward, J.~E. Avila, and E.~Rivas, ``Dialogs {{Re-enacted Across Languages}},'' {University of Texas at El Paso}, Technical UTEP-CS-22-108, 2022.

\bibitem{ward2019}
N.~G. Ward, \emph{Prosodic {{Patterns}} in {{English Conversation}}}.\hskip 1em plus 0.5em minus 0.4em\relax {Cambridge University Press}, 2019.

\bibitem{mary2013}
L.~Mary, A.~Babu K.~K, A.~Joseph, and G.~M. George, ``Evaluation of mimicked speech using prosodic features,'' in \emph{2013 {{IEEE International Conference}} on {{Acoustics}}, {{Speech}} and {{Signal Processing}}}, 2013, pp. 7189--7193.

\bibitem{rilliard2011}
A.~Rilliard, A.~Allauzen, and P.~B. de~Mare{\"u}il, ``Using {{Dynamic Time Warping}} to {{Compute Prosodic Similarity Measures}},'' in \emph{Interspeech}, 2011.

\bibitem{openai2023}
\BIBentryALTinterwordspacing
OpenAI, ``Whisper,'' 2023. [Online]. Available: \url{https://github.com/openai/whisper}
\BIBentrySTDinterwordspacing

\bibitem{ward2022two}
N.~Ward, A.~Kirkland, M.~Wlodarczak, and {\'E}.~Sz{\'e}kely, ``Two pragmatic functions of breathy voice in {A}merican {E}nglish conversation,'' in \emph{11th International Conference on Speech Prosody}, 2022, pp. 82--86.

\bibitem{ward2017}
N.~G. Ward and P.~Gallardo, ``Non-{{Native Differences}} in {{Prosodic-Construction Use}},'' \emph{Dialogue \& Discourse}, vol.~8, no.~1, pp. 1--30, 2017.

\end{thebibliography}

\end{document}